\documentclass[runningheads]{llncs}
\usepackage[T1]{fontenc}
\usepackage{graphicx}
\usepackage{subcaption}
\usepackage{booktabs}
\usepackage{amsmath}
\usepackage{amssymb}
\usepackage{array}
\usepackage{hyperref}
\usepackage{multirow}
\usepackage{xcolor}
\usepackage{enumitem}
\usepackage{placeins}
\usepackage{booktabs}
\usepackage{makecell}
\usepackage{caption}
\begin{document}

\title{MultiAttenGastro: Multi-Dimensional Attention Augmentation for 
Gastrointestinal Endoscopy Classification}

\titlerunning{Multi-Dimensional Attention for GI Endoscopy Classification}

\author{Sadhana Devarajan\inst{1},
        Praveen Kumar Chandaliya\inst{1}, Dhruvin Jashvant Kumar Shah\inst{2}, Kishor Upla\inst{1}, \and Kiran Raja\inst{3}}

\authorrunning{Devarajan et. al.}

\institute{
Sardar Vallabhbhai National Institute of Technology, Surat, Gujarat, India\\
\email{u23ai003@svnit.ac.in, pkc@aid.svnit.ac.in, kpu@eced.svnit.ac.in}
\and
Swasthyam Gastro and Liver Hospital, Surat, Gujarat, India\\
\email{dhruvin.mge@gmail.com}
\and
Norwegian University of Science and Technology, Norway\\
\email{kiran.raja@ntnu.no}
}

\maketitle
\hypersetup{pageanchor=false}

% =============================================================================
\begin{abstract}
Automated gastrointestinal (GI) endoscopy classification requires models that
generalize across diverse modalities and class distributions, often far from
natural-image pretraining. We propose MultiAttenGastro, a plug-and-play
attention framework with parallel 1-D channel, 2-D spatial, and 3-D contextual
heads, and present the first systematic cross-dataset evaluation across eight
CNN and transformer backbones on five public GI datasets (80 backbone--dataset
runs). We find that attention effectiveness is not universal but tracks the
representational gap between ImageNet features and the target distribution:
MultiAttenGastro improves 6 of 8 backbones on Kvasir-Capsule (14-class WCE,
large gap; best macro F1 98.33\%), is uniformly negative on the small-gap
Kvasir-v2 benchmark (0/8), and shows mixed outcomes on datasets with
intermediate gap. Five-seed ablation on the strongest case (Kvasir-Capsule,
ConvNeXt-Tiny) shows this improvement is directionally consistent but not
statistically decisive (paired $t$: $p=0.47$; Wilcoxon: $p=0.63$), and that
individual attention heads are not uniformly beneficial in isolation --- only
their combination yields a positive mean effect. Centered Kernel Alignment
(CKA) analysis links this pattern to representational redundancy: low
inter-head CKA under large domain gaps coincides with the framework's only
consistent gains, while high redundancy under small gaps coincides with its
losses. We report these results, including the non-significant margins, as
evidence for when and why multi-dimensional attention helps GI endoscopy
classification, rather than as a claim that MultiAttenGastro is a strictly
superior architectural choice.
\keywords{Wireless Capsule Endoscopy \and Gastrointestinal Classification \and
Attention Mechanism \and Explainable AI}
\end{abstract}

% =============================================================================
\section{Introduction}
\label{sec:intro}

Wireless capsule endoscopy (WCE) produces up to $50{,}000$ frames over $8$--$10$ hours as a pill-sized camera traverses the GI tract~\cite{muhammad2020wce,iddan2000wireless}. Pathologies such as bleeding, polyps, or erosions may appear in fewer than $0.1\%$ of frames, making automated classification a clinical necessity. The Kvasir-Capsule benchmark~\cite{smedsrud2021kvasir} exemplifies this challenge with $47{,}238$ labelled WCE images across $14$ categories and an extreme $3{,}434{:}1$ imbalance between normal and rare pathology. Gastroscopy datasets such as Kvasir-v2, HyperKvasir, and GastroVision extend the task to $8$--$27$ classes in a modality visually closer to natural images.

We propose \textbf{MultiAttenGastro}, a plug-and-play framework that applies parallel \(1\text{D} + 2\text{D} + 3\text{D}\) attention mechanisms to GI endoscopy classification, capturing channel, spatial, and contextual dependencies for fine-grained lesion discrimination. Unlike prior single-dataset studies, we conduct the first systematic cross-dataset evaluation of multi-dimensional attention across diverse GI imaging conditions. Our findings show that \textbf{attention effectiveness is not universal but
correlates with a measurable factor}: initial results suggested a
modality-dependent effect (helping WCE, hurting gastroscopy), but adding
SEE-AI revealed mixed outcomes that modality alone could not explain. The
strongest associated factor is the \emph{representation gap} between
ImageNet-pretrained features and the target dataset --- large for
Kvasir-Capsule's rare pathologies, small for gastroscopy's natural-image-like
appearance, and small for SEE-AI's simpler four-class taxonomy despite its WCE
modality. We treat this as an empirical association supported by five-seed
ablation, CKA analysis, and Grad-CAM (Sects.~\ref{sec:ablation}--\ref{sec:discussion}),
not as a proven causal mechanism; Sect.~\ref{sec:backbone-cka} shows the
association is not perfect even within our own data.

This paper makes the following contributions:
\begin{enumerate}[label=(\roman*)]
    \item We propose \textit{MultiAttenGastro} and evaluate it across eight
    CNN and transformer backbones and five datasets (80 runs), constituting
    the first cross-dataset study of multi-dimensional attention in GI
    imaging.
    \item We show that MultiAttenGastro's effectiveness correlates with the
    representational domain gap between ImageNet pretraining and the target
    dataset, rather than imaging modality: it helps under large gap
    (Kvasir-Capsule), hurts uniformly under small gap (Kvasir-v2), and is
    mixed at intermediate gap, with SEE-AI (WCE, small gap) confirming that
    modality alone does not determine outcome.
    \item Through five-seed ablation, paired significance testing, and CKA
    analysis, we show that (a) the module's benefit on its strongest case is
    directionally consistent but not statistically significant at $n=5$
    seeds, (b) individual attention heads are not independently beneficial ---
    only their combination is --- and (c) this pattern correlates with
    inter-head representational redundancy, which is itself gap-dependent.
    \item We qualitatively corroborate this framework with Grad-CAM, showing
    lesion-focused activation gains on the high-gap dataset and dispersed,
    non-diagnostic activation shifts on the low-gap dataset.
\end{enumerate}

% =============================================================================
\section{Related Work}\label{sec:related}

\subsection{GI Endoscopy Classification}
Wireless capsule endoscopy (WCE) generates large volumes of redundant imagery, motivating automated analysis~\cite{muhammad2020wce,panchananam2024capsule}. Since the release of Kvasir-Capsule~\cite{smedsrud2021kvasir}, deep learning has advanced GI endoscopy classification considerably. Early work focused on CNN transfer learning: Fonseca et al.~\cite{fonseca2022} used ResNet50 and EfficientNetB3 for binary WCE classification; Li et al.~\cite{li2024} proposed a 17-CNN majority-voting ensemble; AlOtaibi et al.~\cite{alotaibi2024} introduced Efficient-Gastro, an EfficientNet-based transfer-learning framework; and Habe et al.~\cite{habe2025rtdetr} explored transformer-based RT-DETR for real-time WCE lesion detection. Attallah et al.~\cite{attallah2025endonet} proposed EndoNet, combining multi-CNN feature fusion with NNMF dimensionality reduction and mRMR feature selection, reporting 97.8\% accuracy on Kvasir-v2 and 98.4\% on a balanced 10-class HyperKvasir subset. In contrast, MultiAttenGastro is an end-to-end attention-augmented framework evaluated on the complete HyperKvasir taxonomy (19 classes) under natural imbalance—a more clinically realistic assessment.

Table~\ref{tab:sota_comparison} compares our results with these methods on overlapping datasets. Prior work typically reports accuracy on curated single-split subsets, whereas we report macro F1 (and accuracy where available) under stratified multiseed evaluation. Both metrics are shown explicitly rather than implied as equivalent. The RT-DETR result is an object-detection metric and not directly comparable to classification rows.

\begin{table}[t]
\centering
\caption{Comparison with recent GI classification methods on overlapping datasets. Prior work typically reports accuracy on curated subsets, while our results include both accuracy and macro F1 under stratified multiseed evaluation. The RT-DETR entry is an object-detection task and not directly comparable to classification rows.}
\label{tab:sota_comparison}
\scriptsize
\begin{tabular}{lllccc}
\toprule
\textbf{Method} & \textbf{Backbone} & \textbf{Dataset} & \textbf{\#Cls} & \textbf{Acc.} & \textbf{Macro F1} \\
\midrule
\multicolumn{6}{l}{\textit{Kvasir-v2}} \\
Al-Otaibi et al.~\cite{alotaibi2024} & EfficientNet-B1 & Kvasir-v2 & 8 & 98.94\% & -- \\
Attallah et al.~\cite{attallah2025endonet} & CNN fusion $+$ ML & Kvasir-v2 & 8 & 97.8\% & -- \\
MultiAttenGastro & ResNet50 & Kvasir-v2 & 8 & 94.33\% & 94.32\% \\
\midrule
\multicolumn{6}{l}{\textit{HyperKvasir}} \\
Li et al.~\cite{li2024} & 17-CNN ensemble & HyperKvasir  & 10 & 95.30\% & -- \\
&&(10-cls subset)&&&\\
Attallah et al.~\cite{attallah2025endonet} & CNN fusion $+$ ML & HyperKvasir & 10 & 98.4\% & -- \\
&&(10-cls subset)&&&\\
MultiAttenGastro& ViT-B/16 $+$ MAG & HyperKvasir & 19 & 89.93\% & 86.30\% \\
&&(full taxonomy)&&&\\
\midrule
\multicolumn{6}{l}{\textit{Kvasir-Capsule}} \\
Habe et al.~\cite{habe2025rtdetr}  & ResNet-101 & Kvasir-Capsule & 10$+$bg & -- & 98.0\%$^{\dagger}$ \\
(RT-DETR-X) & & & & &\\
MultiAttenGastro& ConvNeXt-Tiny  & Kvasir-Capsule & 14 & 98.77\% & 98.33\% \\
& $+$ MAG & & & &\\
\midrule
\multicolumn{6}{l}{\textit{Other WCE}} \\
Al-Otaibi et al.~\cite{alotaibi2024} & EfficientNet-B1 & WCE (4-class) & 4 & 96.63\% & -- \\
\bottomrule
\multicolumn{6}{l}{\footnotesize $^{\dagger}$Weighted F1 on an object-detection task; not directly comparable to macro F1 above.}
\end{tabular}
\end{table}

On overlapping datasets, MultiAttenGastro achieves strong performance but does not match the highest reported accuracies on curated subsets. For example, on Kvasir-v2 our ResNet50 baseline reaches 94.3\% accuracy, while prior EfficientNet-based and fusion approaches report 97.8--98.9\%. On HyperKvasir, our ViT-B/16 $+$ MAG achieves 89.9\% accuracy and 86.3\% macro F1 on the full 19-class taxonomy, compared to 95.3--98.4\% accuracy on balanced 10-class subsets. On Kvasir-Capsule, MultiAttenGastro with ConvNeXt-Tiny attains 98.8\% accuracy and 98.3\% macro F1, comparable to detection-focused RT-DETR. These results highlight that while absolute accuracy varies with dataset scope, our contribution lies in the first systematic cross-dataset and cross-architecture evaluation of multi-dimensional attention, identifying when augmentation helps or hurts and why.

On same-metric terms, MultiAttenGastro trails prior work by a modest margin
on Kvasir-v2 (94.33\% vs.\ 97.8--98.94\% accuracy) and a larger margin on
HyperKvasir (89.93\% vs.\ 95.30--98.4\% accuracy). We attribute part of the
HyperKvasir gap to task difficulty: prior work evaluates a curated 10-class
subset, whereas we evaluate the complete 19-class taxonomy under natural
imbalance. Matching state-of-the-art accuracy on a single curated split is
not this paper's goal; the paper's contribution is a systematic cross-dataset
and cross-architecture study of \emph{when} multi-dimensional attention helps
or hurts (Sects.~\ref{sec:ablation}--\ref{sec:discussion}). We flag this
distinction explicitly because our evidence for \emph{why} it helps or hurts
is correlational and, on our own multi-seed tests, not statistically decisive
at the single-dataset level (Sect.~\ref{sec:attn-comparison}); readers should
treat the domain-gap account as the best-supported explanation among those we
tested, not as a settled mechanism. Beyond backbone architecture, feature
fusion, or detection performance --- the primary emphases of prior work ---
the generalization of multi-dimensional attention across GI datasets,
backbones, and domain gaps remains largely unexplored; our work addresses
this gap directly, with the caveats stated above.

\subsection{Attention Mechanisms in Medical Imaging}
Attention mechanisms have become central to medical image analysis, enabling networks to highlight diagnostically relevant features while suppressing irrelevant information. Early approaches include Squeeze-and-Excitation (SE) networks~\cite{hu2018squeeze}, which recalibrate channel responses, and CBAM~\cite{woo2018cbam}, which sequentially combines channel and spatial attention. Attention gates have been widely adopted for segmentation~\cite{oktay2018attunet,schlemper2019attgate}, with more recent extensions including AW-Net~\cite{pal2023awnet} and PanNet~\cite{pal2025pannet}.These attention-gate formulations are primarily developed for segmentation,
where gating suppresses irrelevant activations at skip connections to
sharpen localisation of a target region within a U-Net-style encoder-decoder.
MultiAttenGastro targets a different problem and a different point in the
network: rather than gating skip-connection features for pixel-level
localisation, it applies parallel channel, spatial, and contextual attention
to backbone features for image-level classification, with no segmentation
decoder or skip-connection structure involved.

MultiAttenGastro differs from existing modules in both design and objective. Unlike CBAM~\cite{woo2018cbam} and Triplet Attention~\cite{misra2021triplet}, which refine features sequentially, or Coordinate Attention~\cite{hou2021coordattn} and ECA-Net~\cite{wang2020ecanet}, which emphasize channel interactions, MultiAttenGastro computes independent 1-D channel, 2-D spatial, and 3-D contextual branches in parallel and fuses them by element-wise addition. This structure enables direct head-level ablation. Rather than introducing a new attention mechanism, our contribution is the first comprehensive evaluation of when multi-dimensional attention improves or degrades GI endoscopy classification, showing that its effectiveness is governed by the representational domain gap between ImageNet pretraining and the target dataset.

\subsection{Explainable AI in GI Endoscopy}
Explainable AI (XAI) is increasingly important in GI endoscopy, where model transparency is essential for clinical trust and adoption. Despite high performance in deep learning for lesion detection, classification, and segmentation, black-box opacity limits interpretability. Grad-CAM~\cite{selvaraju2017gradcam} has emerged as the most widely used saliency method, with recent work highlighting XAI as a clinical and regulatory necessity~\cite{mascarenhas2025xai} and Nadimi et al.~\cite{nadimi2025xai} validating Grad-CAM within a complete WCE workflow. In this work, Grad-CAM serves not only as an interpretability tool but as mechanistic evidence supporting our domain-gap hypothesis. We employ Grad-CAM in preference to perturbation-based methods such as
Score-CAM~\cite{wang2020scorecam}, LIME~\cite{ribeiro2016lime}, and
SHAP~\cite{lundberg2017shap} for three reasons specific to our setting.
First, Grad-CAM requires a single backward pass per image, which matters at
our experimental scale (80 backbone--dataset runs); Score-CAM's repeated
forward-pass perturbation scheme scales poorly by comparison. Second,
Grad-CAM operates directly on the convolutional feature maps that our
attention module modifies, making it a more direct diagnostic of the
attention heads themselves than LIME or SHAP's superpixel- or
feature-level surrogate approaches, which are primarily validated on
natural-image or tabular data rather than dense medical feature maps.
Third, Grad-CAM has prior validation within complete clinical WCE
workflows~\cite{nadimi2025xai}, giving it more direct precedent in our
target domain. We treat this as a deliberate methodological choice suited
to our setting rather than a claim of categorical superiority; a
gradient-free method such as Score-CAM could offer a useful cross-check in
future work.
% =============================================================================
\begin{figure}[tp]
    \centering
    \includegraphics[width=0.95\linewidth]{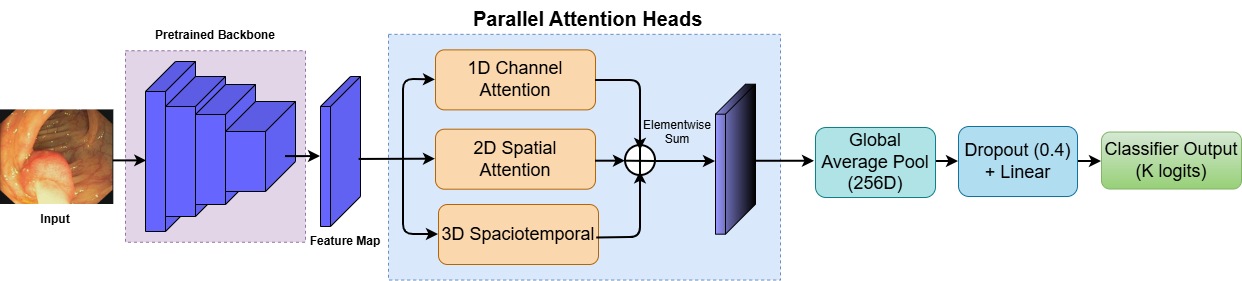}
    \caption{Architecture of the proposed MultiAttenGastro architecture for GI endoscopy classification. A pretrained CNN backbone first extracts deep feature maps, followed by a $1\times1$ convolutional layer that projects the features to a 256-channel representation. The refined features are processed by three parallel attention branches: (i) channel-wise (1D), (ii) spatial (2D), and (iii) contextual (3D). The outputs of all attention heads are fused via element-wise addition ($\oplus$), enabling complementary feature integration. The fused representation is subsequently aggregated using global average pooling (GAP) and passed to a fully connected layer for final class prediction.}
    \label{fig:architecture}
\end{figure}
\section{Proposed Architecture}\label{sec:method}
\subsection{MultiAttenGastro for GI Endoscopy}
MultiAttenGastro is \emph{plug-and-play} and backbone-agnostic: it attaches to the output of any pretrained CNN or transformer feature extractor without modifying the host architecture. While not training-free or zero-shot, the module and, in Phase~2, the top backbone layers are optimized via supervised fine-tuning (Sect.~\ref{sec:data}). Fig.~\ref{fig:architecture} illustrates the pipeline.

A pretrained backbone extracts feature maps from an input endoscopy image. A $1\times1$ convolution with BatchNorm and ReLU projects the output to $C=256$ channels, yielding $X_{in} \in \mathbb{R}^{H\times W\times 256}$. Three parallel attention branches are then applied:

\begin{itemize}
    \item \textbf{1-D Channel Attention}: global average pooling followed by two $1\times1$ convolutions with ReLU and Sigmoid activations generates per-channel weights, amplifying informative features while suppressing less relevant ones.
    \item \textbf{2-D Spatial Attention}: captures cross-dimensional dependencies across $(c,w)$, $(c,h)$, and $(h,w)$ planes, adaptively fused with learnable scalars $\omega_1, \omega_2, \omega_3$, encoding relationships between channel semantics and spatial structure.
    \item \textbf{3-D Contextual Attention}: applies a depthwise convolution (kernel size 3, padding 1) followed by a dilated depthwise convolution (kernel size 3, dilation 3) and pointwise GELU activation, capturing both local and long-range spatial context within a single static image. We use the term ``contextual'' rather than ``spatiotemporal'' to avoid implying sequence modeling.
\end{itemize}

The outputs are fused via element-wise addition, $X_{out} = X_1 \oplus X_2 \oplus X_3$, followed by global average pooling to a 256-dimensional representation, dropout (rate 0.4), and a linear classifier with $K$ output classes. Backbone interaction with the module is not uniform across architectures: some backbones benefit consistently, others show mixed or near-null effects. Sect.~\ref{sec:discussion} quantifies this variation with a four-backbone, five-seed multiseed comparison on Kvasir-Capsule.

% =============================================================================
\section{Datasets and Experimental Setup}\label{sec:data}

\subsection{Datasets}
We evaluate MultiAttenGastro on five publicly available GI endoscopy datasets (Table~\ref{tab:datasets}), spanning two imaging modalities and a wide range of class distributions and visual complexity. \textbf{Kvasir-Capsule} contains 14 rare pathological classes substantially different from the ImageNet distribution, yielding the largest domain gap in our benchmark; its extreme $3{,}434{:}1$ imbalance ratio requires careful preprocessing. \textbf{SEE-AI} is another WCE dataset with four classes (Normal, Ulcer, Polyp, Bleeding); despite sharing Kvasir-Capsule’s modality, its simpler taxonomy reduces the effective domain gap, making it a critical robustness benchmark. \textbf{Kvasir-v2}, \textbf{HyperKvasir}, and \textbf{GastroVision} are gastroscopy datasets with 8--27 classes, visually closer to natural images and thus exhibiting smaller domain shifts from ImageNet.

\begin{table}[tp]
\centering
\caption{Dataset overview with split sizes, domain gap, and MultiAttenGastro win/loss counts across the 8-backbone benchmark (Tables~\ref{tab:kvasirc}, \ref{tab:otherdatasets}).}
\label{tab:datasets}
\label{tab:cross_dataset}
\setlength{\tabcolsep}{4pt}
\scriptsize
\begin{tabular}{llccccccc}
\toprule
\textbf{Dataset} & \textbf{Modality} & \textbf{Train} & \textbf{Val} & \textbf{Test} & \textbf{Imbalance} & \textbf{Gap} & \textbf{Wins} & \textbf{Losses} \\
\midrule
Kvasir-Capsule~\cite{smedsrud2021kvasir} & WCE & 9,299 & 1,641 & 2,354 & 3,434:1 & Large & 6/8 & 2/8 \\
SEE-AI\footnote{\url{https://www.kaggle.com/datasets/capsuleyolo/kyucapsule}} & WCE & 12,943 & 2,769 & 2,769 & Moderate & Small & 4/8 & 4/8 \\
Kvasir-v2~\cite{smedsrud2021kvasir} & Gastroscopy & 5,600 & 1,200 & 1,200 & Balanced & Small & 0/8 & 8/8 \\
HyperKvasir & Gastroscopy & 6,799 & 1,450 & 1,450 & Moderate & Small--Med & 3/8 & 5/8 \\
GastroVision & Gastroscopy & 7,785 & 1,662 & 1,662 & Moderate & Small--Med & 2/8 & 6/8 \\
\bottomrule
\end{tabular}
\end{table}

\subsection{Class Imbalance Handling}
Training directly on the raw Kvasir-Capsule distribution biases models toward the dominant class, an unacceptable outcome that fails to detect rare pathologies. We validate a two-stage correction using EfficientNet-B3 (25 epochs): raw data (F1: 88.69\%), under-sampling only (F1: 95.88\%), and under-sampling $+$ targeted augmentation (F1: 97.65\%)—an 8.96-point gain adopted for all Kvasir-Capsule experiments. \textbf{Under-sampling} caps majority classes at 3,000 samples ($3{,}434{:}1 \rightarrow 300{:}1$), retaining all minority samples. \textbf{Targeted augmentation} then raises the three most underrepresented classes (Polyp $10{\rightarrow}500$, Blood-Hematin $12{\rightarrow}300$, Ampulla of Vater $10{\rightarrow}200$; imbalance $\rightarrow 18.9{:}1$) using flips, $\pm30^\circ$ rotations, colour jitter, and Gaussian noise—consistent with capsule motion artefacts (colour inversion excluded as it distorts tissue characteristics). A \textbf{class-weighted loss} ($w_i = N/(K \cdot n_i)$) further penalises minority-class errors up to $18.8\times$ more than majority-class errors.

\begin{table}[!htbp]
\centering
\scriptsize
\caption{\textbf{Kvasir-Capsule} (WCE, 14 classes, large domain gap), single-seed (seed=42). MultiAttenGastro improves 6 of 8 backbones. See Sect.~\ref{sec:discussion} for multiseed re-evaluation of DenseNet121 and MobileNet-v3-L exceptions.}
\label{tab:kvasirc}
\begin{tabular}{lcccc}
\toprule
\textbf{Backbone} & \textbf{Params(Base)} & \textbf{Params($+$MAG)} & \textbf{F1 Base(\%)} & \textbf{F1$+$MAG($\Delta$ \%)} \\
\midrule
ResNet50        & 23.54M & 24.25M & 97.29 & 97.78 (+0.49) \\
DenseNet121     &  6.97M &  7.43M & 98.04 & 97.80 ($-$0.24) \\
EfficientNet-B3 & 10.72M & 11.30M & 97.33 & 97.68 (+0.35) \\
Inception-v3    & 21.81M & 22.52M & 96.97 & 97.45 (+0.48) \\
MobileNet-v2    &  2.24M &  2.77M & 97.13 & 98.10 (+0.97) \\
MobileNet-v3-L  &  2.99M &  3.43M & 97.79 & 97.49 ($-$0.30) \\
ViT-B/16        & 86.58M & 86.98M & 97.49 & 98.10 (+0.61) \\
ConvNeXt-Tiny   & 27.83M & 28.23M & 97.28 & \textbf{98.33} (+1.05) \\
\bottomrule
\end{tabular}
\end{table}

\subsection{Preprocessing and Training}
All images are resized to $224\times224$ (or $299\times299$ for Inception-v3) and normalized using ImageNet statistics. For Kvasir-Capsule, we follow the official train/test split and allocate 15\% of the training set for validation (seed = 42). HyperKvasir and GastroVision are partitioned using stratified 70/15/15 splits, while SEE-AI and Kvasir-v2 retain their predefined splits (Table~\ref{tab:datasets}). In total, we conduct 80 runs (8 backbones $\times$ 2 conditions $\times$ 5 datasets), each trained for 92 epochs with the Adam optimizer and weight decay $10^{-4}$. Unless otherwise noted (Sect.~\ref{sec:ablation}), a fixed random seed (42) ensures reproducibility of the backbone benchmark. Batch size is set to 16 for Kvasir-Capsule and 32 for the remaining datasets. Hyperparameters are kept consistent across experiments, ensuring that observed differences arise solely from the attention module and dataset characteristics.

\subsection{Two-Phase Training Protocol}
We adopt a two-phase strategy to stabilize attention learning before adapting the backbone:

\begin{itemize}
    \item \textbf{Phase 1 (warm-up, 2 epochs):} the backbone is frozen; only the channel-reduction layers, attention module, and classifier are optimized at $\mathrm{lr}=10^{-3}$.
    \item \textbf{Phase 2 (fine-tuning, 90 epochs):} the top 30\% of backbone layers are unfrozen and jointly trained at $\mathrm{lr}=10^{-4}$, with a ReduceLROnPlateau scheduler (factor $=0.5$, patience $=5$) adaptively lowering the rate on plateaus common in imbalanced training.
\end{itemize}

% =============================================================================
\section{Quantitative Results}\label{sec:quantitative}
\subsection{Backbone Benchmark}
Tables~\ref{tab:kvasirc}--\ref{tab:seeai} report macro F1 across all backbone--dataset combinations. On \textbf{Kvasir-Capsule} (Table~\ref{tab:kvasirc}), MultiAttenGastro improves 6 of 8 backbones, led by ConvNeXt-Tiny at 98.33\% (+1.05\%) and MobileNet-v2 at 98.10\% (+0.97\%). These gains highlight the module’s effectiveness under large domain gaps and extreme imbalance. The two single-seed exceptions, DenseNet121 ($-0.24\%$) and MobileNet-v3-L ($-0.30\%$), are revisited in Sect.~\ref{sec:discussion}, where multiseed re-evaluation shows MobileNet-v3-L reverses to a small positive mean effect.

On \textbf{Kvasir-v2} (Table~\ref{tab:kvasirv2}), MultiAttenGastro consistently reduces performance across all 8 backbones ($-0.14\%$ to $-2.79\%$). This uniformity suggests a dataset-level effect, consistent with gastroscopy’s small domain gap, where additional attention introduces redundancy rather than complementary cues.

\textbf{HyperKvasir} and \textbf{GastroVision} (Table~\ref{tab:otherdatasets}) show mixed outcomes (3/8 and 2/8 wins), aligning with moderate domain gaps. MobileNet-v2 incurs the largest drops ($-3.45\%$, $-3.82\%$), contrasting with its positive behaviour on Kvasir-Capsule, indicating dataset-specific interactions rather than fixed backbone properties.

On \textbf{SEE-AI} (Table~\ref{tab:seeai}; WCE, 4 classes), MultiAttenGastro shows balanced outcomes (4/8 wins), mirroring the gastroscopy datasets rather than Kvasir-Capsule. Together, these results confirm that attention effectiveness is governed by the representational domain gap rather than imaging modality.

\begin{table}[htbp]
\centering
\tiny
\setlength{\tabcolsep}{3pt}
\caption{Backbone benchmark on the four remaining datasets (F1 Base(\%) / $+$MAG($\Delta$\%)). Bold = best F1 in each dataset. Kvasir-v2: 0/8 wins (uniform degradation, small domain gap). HyperKvasir: 3/8 wins (small–medium gap). GastroVision: 2/8 wins (small–medium gap). SEE-AI: 4/8 wins (WCE, small effective gap).}
\label{tab:kvasirv2}
\label{tab:hyperkvasir}
\label{tab:gastrovision}
\label{tab:seeai}
\label{tab:otherdatasets}
\begin{tabular}{l cc cc cc cc}
\toprule
& \multicolumn{2}{c}{\textbf{Kvasir-v2}} & \multicolumn{2}{c}{\textbf{HyperKvasir}} & \multicolumn{2}{c}{\textbf{GastroVision}} & \multicolumn{2}{c}{\textbf{SEE-AI}} \\
\textbf{Backbone} & Base & $+$MAG($\Delta$) & Base & $+$MAG($\Delta$) & Base & $+$MAG($\Delta$) & Base & $+$MAG($\Delta$) \\
\midrule
ResNet50        & \textbf{94.32} & 93.31($-$1.01) & 83.89 & 84.63(+0.74) & 89.09 & 88.07($-$1.02) & 69.61 & 69.81(+0.20) \\
DenseNet121     & 93.20 & 92.79($-$0.41) & 83.75 & 83.48($-$0.27) & 88.48 & 89.00(+0.52) & 75.04 & 66.33($-$8.71) \\
EfficientNet-B3 & 93.74 & 91.87($-$1.87) & 83.82 & 83.00($-$0.82) & 88.71 & 87.82($-$0.89) & 74.19 & 74.24(+0.05) \\
Inception-v3    & 91.98 & 89.19($-$2.79) & 80.71 & 81.69(+0.98) & 87.19 & 85.75($-$1.44) & 62.37 & 58.30($-$4.07) \\
MobileNet-v2    & 93.08 & 92.48($-$0.60) & 84.94 & 81.49($-$3.45) & \textbf{90.45} & 86.63($-$3.82) & 70.44 & 66.30($-$4.14) \\
MobileNet-v3-L  & 92.55 & 92.41($-$0.14) & 83.18 & 84.78(+1.60) & 88.17 & 87.80($-$0.37) & 72.01 & 75.36(+3.35) \\
ViT-B/16        & 92.76 & 91.75($-$1.01) & 85.25 & \textbf{86.30}(+1.05) & 88.84 & 87.33($-$1.51) & 79.14 & 79.57(+0.43) \\
ConvNeXt-Tiny   & 93.50 & 92.76($-$0.74) & 85.93 & 83.72($-$2.21) & 88.44 & 88.58(+0.14) & \textbf{80.16} & 79.55($-$0.61) \\
\bottomrule
\end{tabular}
\end{table}

% =============================================================================
\subsection{Impact of Effective Domain Gap} 

Table~\ref{tab:cross_dataset} (Wins/Losses columns) highlights that the dominant factor in MultiAttenGastro effectiveness is the \emph{effective domain gap}, rather than imaging modality. The model gains substantially (6/8 wins) under a large domain gap (Kvasir-Capsule), suggesting attention learns mutually discriminative features under high variability, while it uniformly loses (0/8) under a small gap (Kvasir-v2), indicating redundancy rather than complementary cues. Intermediate-gap datasets (HyperKvasir, GastroVision) show blended behaviour (3/8, 2/8), dependent on backbone interactions. SEE-AI, despite being WCE, behaves like the gastroscopy datasets (4/8) rather than Kvasir-Capsule (6/8), confirming that modality alone is not determinative—the key driver is visual complexity and representational distance from ImageNet, with modality serving only as a proxy.

\begin{table}[!ht]
\scriptsize
\centering
\caption{Head-level ablation and alternative-mechanism comparison on Kvasir-Capsule (ConvNeXt-Tiny), 5-seed evaluation (seeds: 7, 42, 99, 123, 2024). Top block: per-head ablation (1D and 2D individually trail baseline; only 3D and the full combination reach or exceed it). Bottom block: alternative attention mechanisms substituted for full MAG under the same protocol. Significance columns report paired tests (matched by seed) against Full MAG; none reach $p<0.05$.}
\label{tab:ablation}
\label{tab:attn-comparison}
\begin{tabular}{l c c c c c}
\toprule
\textbf{Configuration} & \textbf{F1$\mu$$\pm$$\sigma$(\%)} & \textbf{AUC(\%)} & \textbf{$\Delta$F1 vs.\ base(\%)} & \textbf{paired $t$ ($p$)} & \textbf{Wilcoxon $p$} \\
\midrule
No Attention & 98.096 $\pm$ 0.341 & 99.973 & --- & 0.793 (0.472) & 0.625 \\
1D Channel        & 97.694 $\pm$ 0.598 & --- & $-$0.402 & --- & --- \\
2D Spatial        & 97.894 $\pm$ 0.323 & --- & $-$0.202 & --- & --- \\
3D Contextual     & 98.132 $\pm$ 0.127 & --- & +0.036 & --- & --- \\
\midrule
SE-Net~\cite{hu2018squeeze}            & 97.990 $\pm$ 0.341 & --- & $-$0.106 & 2.394 (0.075) & 0.125 \\
CBAM~\cite{woo2018cbam}                & 98.098 $\pm$ 0.351 & --- & +0.002 & 0.919 (0.410) & 0.438 \\
ECA-Net~\cite{wang2020ecanet}          & 97.798 $\pm$ 1.042 & --- & $-$0.298 & 0.927 (0.406) & 1.000 \\
Coordinate Attention~\cite{hou2021coordattn} & 98.118 $\pm$ 0.148 & --- & +0.022 & 1.355 (0.247) & 0.375 \\
Triplet Attention~\cite{misra2021triplet}    & 97.764 $\pm$ 0.984 & --- & $-$0.332 & 1.464 (0.217) & 0.188 \\
\midrule
\textbf{Full MAG}     & \textbf{98.262 $\pm$ 0.337} & \textbf{99.930} & \textbf{+0.166} & --- & --- \\
\textbf{(1D+2D+3D)} &&&&&\\
\bottomrule
\end{tabular}
\end{table}

\subsection{Attention Interference Ablation Study}\label{sec:ablation}
Table~\ref{tab:ablation} illustrates how individual attention heads behave on Kvasir-Capsule with ConvNeXt-Tiny under multi-seed evaluation. The 1D and 2D heads fall below baseline ($-$0.402\%, $-$0.202\%), the 3D head is roughly on par (+0.036\%), and only the full three-head combination yields a consistent mean improvement (+0.166\%). This indicates that individual heads are not uniformly beneficial in isolation, and the module’s value emerges specifically from their combination. The behaviour is distinct from destructive interference (combination worse than all individual heads) and naive additivity, and aligns with the low inter-head CKA observed in Sect.~\ref{sec:discussion}. Per-head ablation thus remains a useful diagnostic step when adapting multi-dimensional attention to new domains.

\subsubsection{Comparison against Alternative Attention Mechanisms}\label{sec:attn-comparison}
To contextualise MultiAttenGastro’s per-head behaviour, Table~\ref{tab:attn-comparison} compares it against established single- and dual-branch mechanisms—SE~\cite{hu2018squeeze}, CBAM~\cite{woo2018cbam}, ECA-Net~\cite{wang2020ecanet}, Coordinate Attention~\cite{hou2021coordattn}, and Triplet Attention~\cite{misra2021triplet}—substituted for the full MultiAttenGastro module on ConvNeXt-Tiny/Kvasir-Capsule under the same 5-seed protocol. Under this protocol, MultiAttenGastro (98.262 $\pm$ 0.337\%) achieves the highest mean macro F1, ahead of CBAM (98.098 $\pm$ 0.351\%) and Coordinate Attention (98.118 $\pm$ 0.148\%). However, paired significance testing shows none of these margins reaches conventional thresholds at $n=5$: the largest effect, MAG vs.\ SE-Net, yields $p=0.075$ (paired $t$) and $p=0.125$ (Wilcoxon). Even the headline MAG-vs-baseline margin (+0.166\%) is not significant ($p=0.472$ paired $t$; $p=0.625$ Wilcoxon). We report this explicitly rather than characterising MultiAttenGastro as a confirmed outright winner: the observed mean ranking is consistent with the data, but margins fall within plausible seed variance, and a larger-$n$ replication would be required for statistical decisiveness. Earlier single-seed comparisons that suggested SE-Net or CBAM outperformed MultiAttenGastro used a different implementation of the attention-comparison pipeline than the one used here; the results reported in Table~\ref{tab:attn-comparison} reflect the 5-seed protocol described above and are the basis for all claims in this paper.

\begin{table}[!htbp]
\centering
\caption{Per-backbone training-time overhead of MultiAttenGastro relative to the no-attention baseline, mean over 5 seeds (7, 42, 99, 123, 2024), Kvasir-Capsule.}
\label{tab:runtime}
\begin{tabular}{l c c c}
\toprule
\textbf{Backbone} & \textbf{Baseline (min)} & \textbf{$+$MAG (min)} & \textbf{Overhead (\%)} \\
\midrule
ConvNeXt-Tiny      & 22.22 & 23.18 & +4.3 \\
MobileNet-v2       & 12.54 & 13.14 & +4.8 \\
DenseNet121        & 24.42 & 26.66 & +9.2 \\
MobileNet-v3-Large & 12.88 & 14.26 & +10.7 \\
\bottomrule
\end{tabular}
\end{table}

Table~\ref{tab:runtime} shows the additional training time introduced by MultiAttenGastro relative to each backbone’s baseline, averaged over five seeds. Overhead ranges from +4.3\% (ConvNeXt-Tiny) to +10.7\% (MobileNet-v3-Large), representing a modest fraction of total training time across backbones of varying capacity.

\subsubsection{Multiseed Evaluation on Extension Datasets}\label{sec:dataset-extension}
To assess consistency beyond Kvasir-Capsule, we also evaluated ConvNeXt-Tiny with and without MultiAttenGastro on Kvasir-v2, HyperKvasir, and SEE-AI across the same five seeds (7, 42, 99, 123, 2024). Results are reported in Table~\ref{tab:dataset-ext}.

\begin{table}[!htbp]
\centering
\scriptsize
\caption{Multiseed evaluation of ConvNeXt-Tiny with and without MultiAttenGastro on Kvasir-v2, HyperKvasir, and SEE-AI (5 seeds each: 7, 42, 99, 123, 2024).}
\label{tab:dataset-ext}
\begin{tabular}{l c c c}
\toprule
\textbf{Dataset} & \textbf{Baseline F1 mean $\pm$ std (\%)} & \textbf{Full MAG F1 mean $\pm$ std (\%)} & \textbf{$\Delta$F1 (\%)} \\
\midrule
Kvasir-v2    & 92.73 $\pm$ 0.471 & 92.56 $\pm$ 0.508 & $-$0.17 \\
HyperKvasir  & 85.77 $\pm$ 0.933 & 86.25 $\pm$ 0.910 & +0.48 \\
SEE-AI       & 77.39 $\pm$ 2.622 & 76.62 $\pm$ 1.685 & $-$0.77 \\
\bottomrule
\end{tabular}
\end{table}

The multiseed averages support the broader domain-gap framework. Kvasir-v2 shows a small negative margin ($-$0.17\%), consistent with redundancy under low gap. HyperKvasir yields a modest positive margin (+0.48\%), reflecting mixed behaviour under intermediate gap. SEE-AI exhibits the largest variance (std.\ 2.622), consistent with its small, heterogeneous class structure; its mean margin is negative ($-$0.77\%), but variability suggests outcomes are class-dependent rather than uniform. Overall, these results reinforce that attention effectiveness is governed by domain gap and dataset-specific representational difficulty.

\begin{figure*}[!htbp]
    \centering
    \begin{subfigure}[t]{0.49\textwidth}
        \centering
        \includegraphics[width=\linewidth]{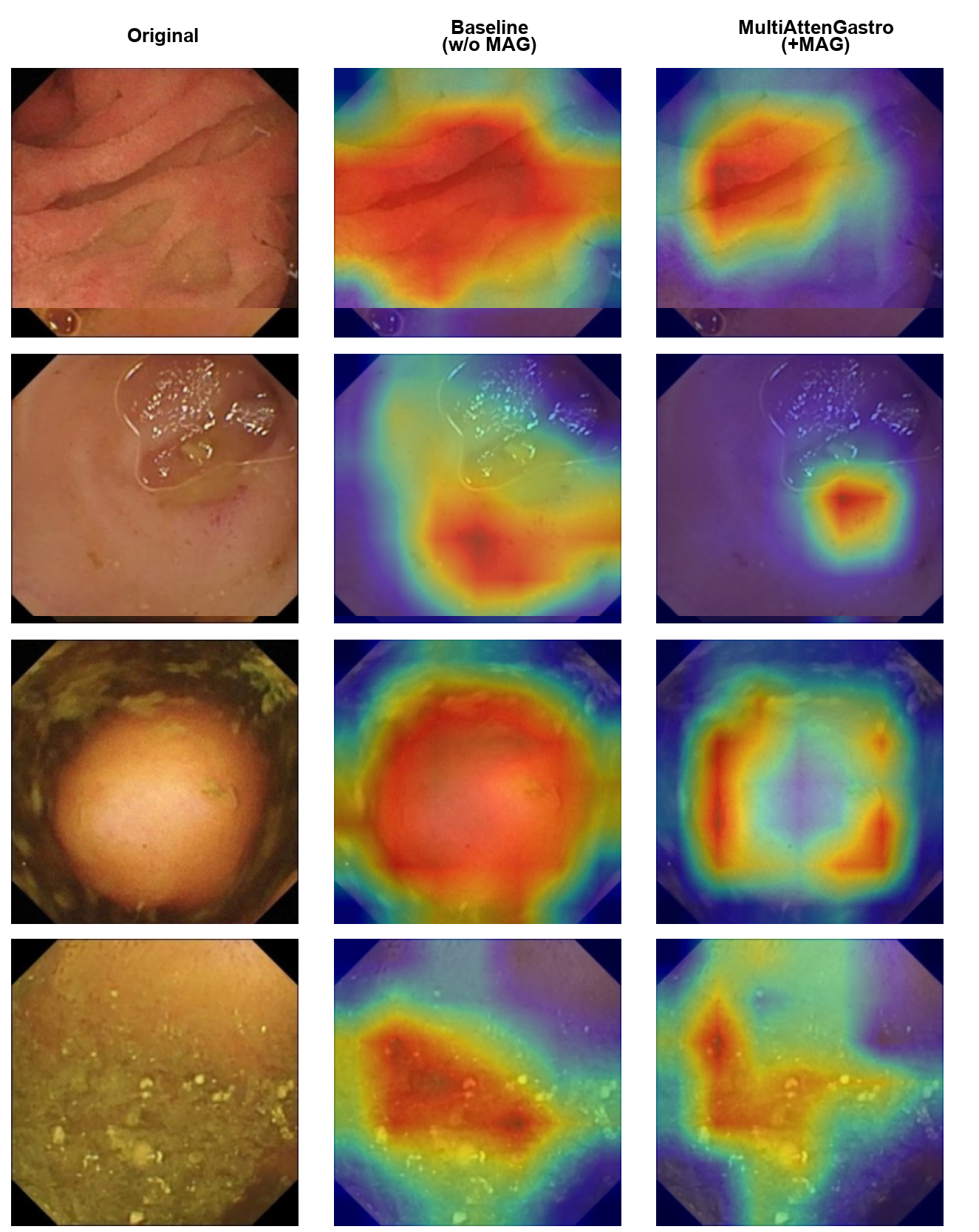}
        \caption{\textbf{Kvasir-Capsule} (WCE; large domain gap). MultiAttenGastro (+MAG) produces sharper lesion-localized activations than the baseline, consistent with performance improvements on 6/8 backbones (Table~\ref{tab:kvasirc}).}
        \label{fig:gradcam_kc}
    \end{subfigure}
    \hfill
    \begin{subfigure}[t]{0.49\textwidth}
        \centering
        \includegraphics[width=\linewidth]{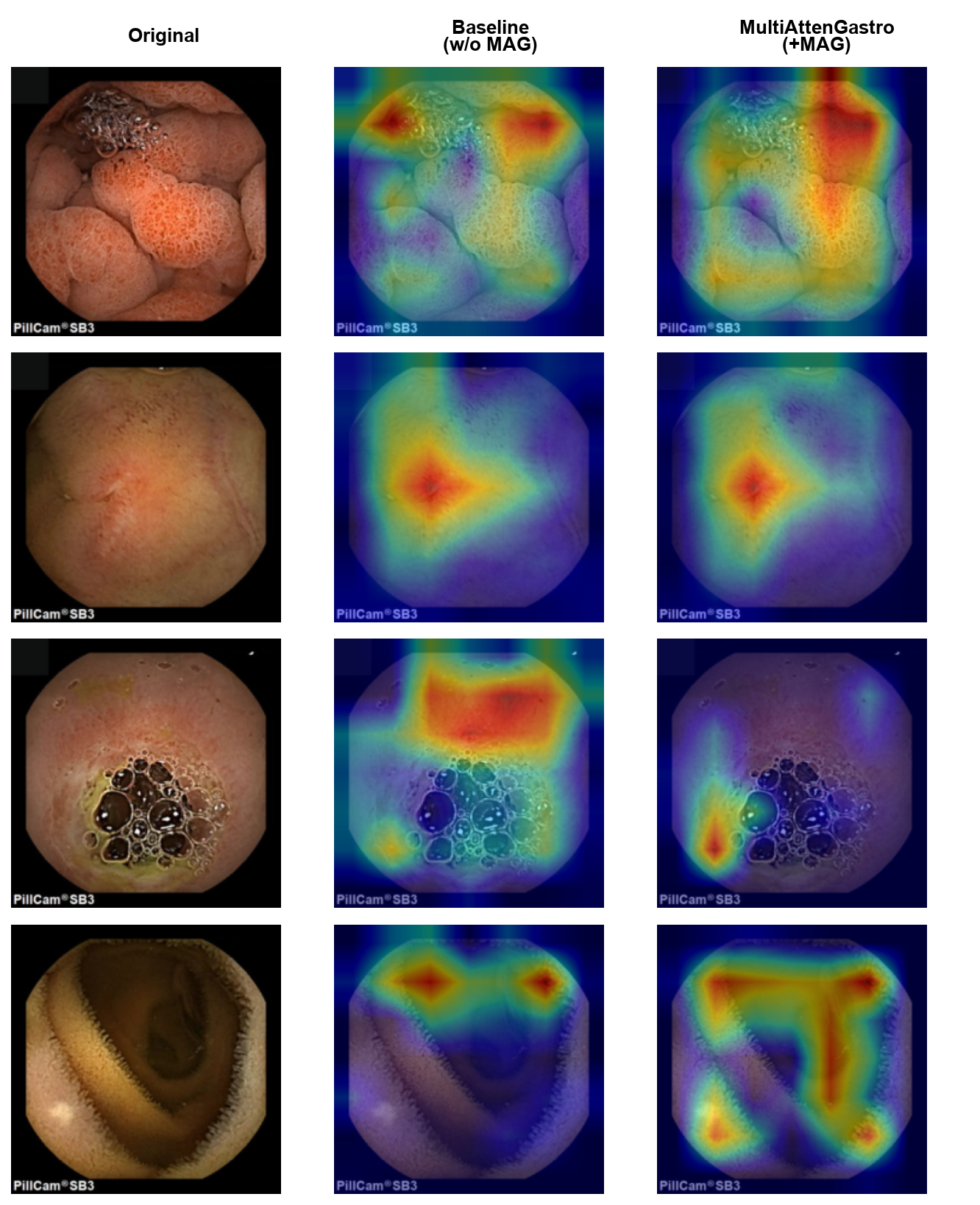}
        \caption{\textbf{SEE-AI} (WCE; small domain gap). +MAG provides clearer localization only for the Bleeding class, while other classes show little or no improvement, matching the quantitative results (Table~\ref{tab:seeai}).}
        \label{fig:gradcam_seeai}
    \end{subfigure}

    \caption{\textbf{Grad-CAM comparison of MultiAttenGastro across high- and low-domain-gap WCE datasets.} Both figures use the same column order: \textit{Original} $\mid$ \textit{Baseline (w/o MAG)} $\mid$ \textit{MultiAttenGastro (+MAG)} with ConvNeXt-Tiny.}
    \label{fig:gradcam_comparison}
\end{figure*}

\section{Qualitative Results}\label{sec:qualitative} 
Grad-CAM visualisations (Figs.~\ref{fig:gradcam_kc}--\ref{fig:gradcam_seeai}), together with the qualitative discussion below, illustrate the domain-gap framework across three distinct behaviours. \textbf{Kvasir-Capsule (large gap):} MultiAttenGastro produces sharply localized activations over diagnostically relevant regions—polyps, bleeding sites, angiectasia—while baselines remain diffuse, consistent with the strong 6/8 backbone improvements (Table~\ref{tab:kvasirc}). \textbf{Kvasir-v2 (small gap):} the trend reverses. Baselines already localize precisely on landmarks (e.g., pylorus) due to strong ImageNet transfer; MultiAttenGastro instead disperses activation across non-diagnostic regions, visually matching the uniform $-0.14\%$ to $-2.79\%$ drops (Table~\ref{tab:kvasirv2}) and reflecting redundancy when heads compete over an already well-aligned representation. \textbf{SEE-AI:} outcomes vary by class. MultiAttenGastro sharpens activation for \emph{Bleeding}, while normal and structural classes show little gain, reinforcing that performance depends on \emph{per-class representational difficulty} rather than modality.

HyperKvasir and GastroVision exhibit diffuse, mixed patterns consistent with their small-to-medium domain gaps and largely negative quantitative outcomes (3/8 and 2/8 wins; Tables~\ref{tab:hyperkvasir},~\ref{tab:gastrovision}). These visualizations are omitted for brevity as they add no qualitatively new insight beyond the cases presented.

% =============================================================================
\section{Discussion}\label{sec:discussion}
MultiAttenGastro is most effective when a substantial representational gap exists between ImageNet pretraining and the target domain—manifesting as weaker baseline performance or pathologies visually unlike natural images (rare lesions, atypical illumination, capsule optics). Conversely, when baselines already generalise well, additional attention can introduce redundancy or conflicting signals, reducing gains. As noted in Sect.~\ref{sec:attn-comparison}, these mean-level trends are not individually significant at $n=5$ seeds and should be read as consistent directional patterns rather than definitive per-comparison effects.

\subsection{Mechanistic Evidence: Representational Redundancy via CKA}
Our initial explanation for destructive-interference results invoked competing gradient signals among heads. A more direct probe is representational redundancy, quantified via centred kernel alignment (CKA). Table~\ref{tab:cka} reports mean inter-head CKA and Grad-CAM insertion-AUC faithfulness gain across all five datasets. Results show higher redundancy (CKA $\geq$0.85) in small-gap datasets, coinciding with negative faithfulness gains, while lower redundancy (CKA $\approx$0.58) in Kvasir-Capsule aligns with positive gains. This supports the domain-gap framework: attention heads contribute complementary features under large gaps but overlap under small gaps, producing interference.
\begin{table}[htbp]
\scriptsize
\centering
\caption{Mean inter-head CKA (representational redundancy) and Grad-CAM faithfulness gain (insertion AUC, MAG vs.\ baseline) across five datasets, ConvNeXt-Tiny.}
\label{tab:cka}
\begin{tabular}{l c c c}
\toprule
\textbf{Dataset} & \textbf{Domain Gap} & \textbf{Mean inter-head CKA} & \textbf{Faithfulness gain} \\
\midrule
Kvasir-Capsule & Large       & 0.577 & +0.021 \\
SEE-AI         & Small       & 0.749 & +0.008 \\
HyperKvasir    & Small-Med   & 0.858 & $-$0.080 \\
GastroVision   & Small-Med   & 0.858 & $-$0.052 \\
Kvasir-v2      & Small       & 0.864 & $-$0.019 \\
\bottomrule
\end{tabular}
\end{table}

The pattern is monotonic: low inter-head CKA (complementary, non-redundant heads) co-occurs with large domain gaps and positive faithfulness gains, while high CKA (redundant heads) aligns with small gaps and negative gains. Kvasir-Capsule stands out with the lowest CKA and the only strongly positive faithfulness gain, matching its status as the dataset where MultiAttenGastro’s benefit is most consistent. This provides mechanistic support for the domain-gap framework: low inter-head CKA is compatible with individual heads contributing little in isolation while their combination captures complementary signal. Under large gaps, the three heads specialise in distinct aspects of the mismatch; under small gaps, an already well-aligned backbone leaves little complementary signal, so heads converge toward redundancy. CKA measures representational overlap, related to but not identical to gradient conflict; we present this as evidence of redundancy rather than direct gradient interference.

\subsection{A Complementary Metric: Backbone Adaptation Magnitude}
\label{sec:backbone-cka}
As a further probe of domain gap, we compute linear CKA~\cite{kornblith2019similarity} between frozen ImageNet-pretrained ConvNeXt-Tiny features and each dataset’s fine-tuned baseline backbone, measuring how far the backbone shifts from ImageNet during training (low CKA = larger shift). Table~\ref{tab:cka_backbone} reports results across all five datasets.

\begin{table}[htbp]
\centering
\scriptsize
\caption{Linear CKA between frozen ImageNet and fine-tuned baseline ConvNeXt-Tiny features. Lower CKA indicates greater feature shift during fine-tuning.}
\label{tab:cka_backbone}
\begin{tabular}{l c c}
\toprule
\textbf{Dataset} & \textbf{CKA(frozen, fine-tuned)} & \textbf{Test $n$} \\
\midrule
SEE-AI          & 0.122 & 2000 \\
Kvasir-Capsule  & 0.225 & 2000 \\
HyperKvasir     & 0.420 & 1450 \\
Kvasir-v2       & 0.459 & 1200 \\
GastroVision    & 0.470 & 1662 \\
\bottomrule
\end{tabular}
\end{table}

This metric agrees with our qualitative labelling for Kvasir-Capsule, Kvasir-v2, HyperKvasir, and GastroVision, but diverges for SEE-AI, whose backbone shows the largest feature shift despite its “Small” qualitative label and mixed (4/8) F1 outcome. We interpret this as evidence that backbone-adaptation magnitude and our qualitative domain-gap construct are related but not interchangeable: SEE-AI’s small class count, moderate imbalance, and constrained training set may independently drive substantial backbone reorganisation. We report this divergence directly, noting it as a limitation of any single proxy; reconciling the two is a useful direction for future work.

\subsection{Backbone-Dependent Interaction, Revisited}
Our per-head ablation (Sect.~\ref{sec:ablation}) and CKA analysis challenge the assumption that adding more attention heads is inherently beneficial, motivating evaluation per backbone and dataset rather than relying on architectural priors. A broader multiseed check across four backbones on Kvasir-Capsule shows a heterogeneous picture: ConvNeXt-Tiny improves in 5/5 seeds (mean $+$1.07\%), MobileNet-v2 in 5/5 seeds ($+$1.28\%), MobileNet-v3-L in 4/5 seeds ($+$0.26\%, reversing the single-seed sign in Table~\ref{tab:kvasirc}), and DenseNet121 is genuinely mixed (3/5 vs.\ 2/5, mean $-$0.05\%). MobileNet-v3-L, once our flagship example of SE-block conflict, in fact benefits on average under multiseed evaluation. We therefore replace the earlier uniform “SE-block conflict” narrative with a backbone-dependent pattern—two consistent gainers, one misleading single-seed case, and one true null—and recommend future compatibility claims be evaluated across multiple seeds before attribution to specific architectural mechanisms.

\textbf{Backbone selection is crucial:} ConvNeXt-Tiny and MobileNet-v2 provide the most consistent and favourable performance–efficiency trade-offs for MultiAttenGastro across all five seeds and are our recommended backbones. MobileNet-v3-L and DenseNet121 exhibit backbone-dependent behaviour and should be validated for each deployment.

\section{Conclusion}\label{sec:conclusion}
We introduced \textbf{MultiAttenGastro}, a plug-and-play multi-dimensional
attention framework for GI endoscopy classification, and used it as a probe
for a broader question: when does attention augmentation help GI
classification, and why. Across 80 backbone--dataset runs, effectiveness
tracks the representational gap between ImageNet pretraining and the target
domain rather than being universally beneficial. The clearest gains occur on
the high-domain-gap Kvasir-Capsule dataset (98.33\% macro F1 with
ConvNeXt-Tiny), while performance declines uniformly on the low-gap Kvasir-v2
benchmark; HyperKvasir, GastroVision, and SEE-AI fall between these extremes.
Five-seed evaluation on the strongest case shows this pattern is directionally
consistent but not statistically decisive ($p=0.47$ paired $t$; $p=0.63$
Wilcoxon), and that the module's benefit arises specifically from combining
three attention heads that are not individually beneficial. CKA analysis
offers a mechanistic account of this behaviour: low inter-head redundancy
under large domain gaps coincides with the framework's only consistent gains,
while high redundancy under small gaps coincides with its losses. Grad-CAM
visualizations corroborate this pattern qualitatively. We present
MultiAttenGastro not as a demonstrated best-in-class attention module ---
our own multi-seed comparison could not distinguish it with statistical
confidence from CBAM or Coordinate Attention on the one dataset where it
performs best (Sect.~\ref{sec:attn-comparison}) but as a concrete case
study and a diagnostic protocol (multi-seed ablation, CKA, significance
testing) for identifying when multi-dimensional attention is likely to help
GI endoscopy models, and when it is likely to add redundant capacity instead.
Our five-seed head-level ablation (Sect.~\ref{sec:ablation}) is currently
reported for a single backbone--dataset pair (ConvNeXt-Tiny, Kvasir-Capsule);
extending this per-head diagnostic across the remaining backbones and
datasets is an important direction for future work.

\section*{Acknowledgements}
The authors acknowledge the Department of AI and CSE,
SVNIT Surat, for providing computational resources on the Dual NVIDIA
H100NVL GPU cluster.

% =============================================================================
\bibliographystyle{splncs04}

\end{document}